\documentclass[conference]{IEEEtran}
\usepackage{array}

\usepackage[utf8]{inputenc}

\usepackage[inline]{enumitem}
\usepackage[normalem]{ulem}
\usepackage[skins,breakable]{tcolorbox}
\usepackage{amsmath,amsfonts}
\usepackage{multirow}
\usepackage{microtype}
\usepackage{framed}
\usepackage{lipsum}
\usepackage{listings}
\usepackage{hyperref}
\usepackage[capitalize]{cleveref}
\crefname{section}{Sect.}{sections}
\Crefname{section}{Section}{Sections}
\usepackage{longtable}
\usepackage{mathtools}
\usepackage{pbox}
\usepackage{relsize}
\usepackage{rotating}
\usepackage{tabularx}
\usepackage{todonotes}
\usepackage{url}
\usepackage{subcaption}
\usepackage{graphicx}
\usepackage{xspace}
\usepackage{tikz}
\usepackage{booktabs}
\usepackage{makecell}
\usetikzlibrary{shapes.geometric}
\usetikzlibrary{arrows, shapes}
\usepackage{float}

\definecolor{xtextBlue}{RGB}{42,8,255}
\newcommand{\bts}{behavior trees\xspace}
\newcommand{\BTCPP}{\texttt{Behavior\-Tree.CPP}\xspace}

\usepackage{ifthen}
\usepackage{amssymb}
\newboolean{showcomments}
\setboolean{showcomments}{true} 
\ifthenelse{\boolean{showcomments}}
{\newcommand{\nb}[2]{
		\fcolorbox{gray}{yellow}{\bfseries\sffamily\scriptsize#1}
		{\sf\small$\blacktriangleright$\textit{#2}$\blacktriangleleft$}
	}
	
}
{\newcommand{\nb}[2]{}
	
}

\definecolor{blue(ncs)}{rgb}{0.0, 0.53, 0.74}
\def\BibTeX{{\rm B\kern-.05em{\sc i\kern-.025em b}\kern-.08em
		T\kern-.1667em\lower.7ex\hbox{E}\kern-.125emX}}

\makeatletter
\newcommand{\linebreakand}{%
\end{@IEEEauthorhalign}
\hfill\mbox{}\par
\mbox{}\hfill\begin{@IEEEauthorhalign}
}
\makeatother

\definecolor{backcolour}{rgb}{0.95,0.95,0.92}
\definecolor{blue-violet}{rgb}{0.54, 0.17, 0.89}

\lstdefinestyle{xmlstyle}{
	language=XML,
	columns=fullflexible,
	backgroundcolor=\color{backcolour},
	keywordstyle=\color{red},
	stringstyle=\color{blue-violet},
	morekeywords={
		name,command,_description,_successIf,_failureIf,_onFailure,action_name},
	classoffset=1,
	morekeywords={False},
	keywordstyle=\color{blue},
	morekeywords={
		BehaviorTreeAction,BehaviorTree,SequenceStar,Sequence,Script,Retry, Fallback,ScriptCondition,Inverter,SubTree}
}

\newcounter{FindingIdx}
\newenvironment{finding}%
{\begin{leftbar}
		\refstepcounter{FindingIdx}
		\noindent\textsc{Finding\,\theFindingIdx.}\@ }%
	{\end{leftbar}}

\begin{document}

\onecolumn
This work has been submitted to the IEEE for possible publication. Copyright may be transferred without notice, after which this version may no longer be accessible.
This version of the contribution has been accepted for
publication at ETFA25 after peer review, but is not the
Version of Record and does not reflect post-acceptance
improvements, or any corrections.

\newpage

\twocolumn
\title{Using Behavior Trees in Risk Assessment}

\author{\IEEEauthorblockN{Razan Ghzouli\IEEEauthorrefmark{1},
		 Atieh Hanna\IEEEauthorrefmark{2},
		Endre Erös\IEEEauthorrefmark{3}, and Rebekka Wohlrab\IEEEauthorrefmark{1}\IEEEauthorrefmark{4}}
	\IEEEauthorblockA{
		\IEEEauthorrefmark{1}Chalmers University of Technology and University of Gothenburg, Gothenburg, Sweden}	
	\IEEEauthorblockA{
		\IEEEauthorrefmark{2}Volvo Group,
		Gothenburg, Sweden}
	\IEEEauthorblockA{
		\IEEEauthorrefmark{3}Chalmers Industriteknik,
		Gothenburg, Sweden}
	\IEEEauthorblockA{
		\IEEEauthorrefmark{4}Carnegie Mellon University, Pittsburgh, USA}
	}

\maketitle

\begin{abstract}
Cyber-physical production systems increasingly involve collaborative robotic missions, requiring more demand for robust and safe missions. Industries rely on risk assessments to identify potential failures and implement measures to mitigate their risks. Although it is recommended to conduct risk assessments early in the design of robotic missions, the state of practice in the industry is different. Safety experts often struggle to completely understand robotics missions at the early design stages of projects and to ensure that the output of risk assessments is adequately considered during implementation.  

\looseness -1 This paper presents a design science study that conceived a model-based approach for early risk assessment in a development-centric way. Our approach supports risk assessment activities by using the behavior-tree model. 
We evaluated the approach together with five practitioners from four companies.
Our findings highlight the potential of the behavior-tree model in supporting early identification, visualisation, and bridging the gap between code implementation and risk assessments' outputs. This approach is the first attempt to use the behavior-tree model to support risk assessment; thus, the findings highlight the need for further development.

\end{abstract}

%



\section{Introduction}
\label{sec:intro}

\looseness=-1
Cyber-Physical Production Systems (CPPSs) increasingly involve collaborative settings where humans and robots work side by side.
Safety and robustness are crucial properties that need to be guaranteed in robotic missions and systems, especially in collaborative missions.
In a large-scale empirical study on the state of robotics software engineering, the robustness of robotics systems was ranked as the most pressing challenge in practice~\cite{garcia2020robotics}.

Risk assessment is the first step of safety analysis, which aims to identify different hazards that can lead to system failure~\cite{guiochet2017safety}. In the context of robotic missions, when an action fails, it can lead to reduced performance or even more catastrophic safety risks. 
Not capturing failures early in the development of robotics systems may lead to technical debt, making future modifications and maintenance more complex and expensive \cite{tom2013exploration}.
However, as with any non-code artifacts or models, there is a risk that practitioners do not see the value of risk assessments if they are disconnected from the actual implementation of robotic missions~\cite{foidl2019technical}. To facilitate their practical adoption, it is crucial to ensure that the outputs of risk assessments are well-integrated and transferred into the code implementation of robotics missions. 
Currently, there is a gap between safety analysis and implementation artifacts, making it difficult for practitioners to see the value of risk assessments~\cite{curtis2012risk}. This paper aims to address these challenges by presenting a model-based approach that supports early risk assessment in a development-centric way.

In the software engineering community, various modeling approaches have been conceived to support the development of robotic missions and systems.
While modeling approaches are not always adopted in practice, one approach that has gained increasing popularity among robotics practitioners is behavior trees. 
Behavior trees model and execute the actions and control-switching mechanisms involved in a robotic mission. 
Key benefits of behavior trees are their understandability~\cite{marzinotto2014towards} and their support to account for reactiveness, by modeling fallback behaviors in case an action cannot be successfully completed~\cite{colledanchise2018behavior}.
Due to their popularity, it appears promising to investigate the use of behavior trees for supporting risk assessment.
To the best of our knowledge, there is no previous research on using behavior trees for safety risk assessment.

\looseness -1 This paper presents a design science study that aims to conceive a model-based approach for early risk assessment in a development-centric way.
We iteratively aimed to understand current problems, develop the model-based risk assessment approach, and evaluate it with five practitioners.
Our approach integrates behavior trees into the risk assessment process to overcome practical challenges. 
We found that using an explicit mission model at the early design stages of the risk assessment can enhance comprehension of the mission and identification of possible failures. Our findings highlight the potential of behavior trees for holding and documenting the outputs of the risk assessment used, as well as aligning the implementation code with assessment outputs. Our findings further demonstrate the challenge of finding the right granularity level of the behavior-tree model, and the necessity of an automated approach and improved tooling to transfer assessment outputs. We see our findings as a promising first step, opening up new directions for future research in using behavior trees in risk assessments. All our data is available at an
accompanying online appendix \cite{appendix:online}.

\section{Background}
\label{sec:background}

\subsection{Behavior Trees}
Behavior trees are modular and flexible, providing intuitive and understandable models for robotics missions \cite{colledanchise2018behavior}.
A behavior-tree model is a directed tree structure containing a root node, control-flow nodes (non-leaf nodes), and execution nodes (leaf nodes). Execution proceeds via "ticks" originating from the root and traverses the tree based on the control-flow node semantics.
Control-flow nodes orchestrate the execution of child nodes, enabling complex task coordination. The four main types of control-flow nodes are sequence, selector, decorator, and parallel nodes. Execution nodes are where the actual execution of the code is defined. They are either robotic actions or conditions evaluating propositions. Each ticked node returns a status (success, failure, or running) to its parent, facilitating dynamic behavior adaptation.

\BTCPP\footnote{https://github.com/BehaviorTree/BehaviorTree.CPP} is the C++ implementation of \bts, and one of the most used libraries in open-source projects \cite{ghzouli2023behavior}. 
\BTCPP has a graphical user interface called Groot that supports the creation and monitoring of \bts. We direct interested readers to previous works for more information on available languages for behavior trees \cite{ghzouli2023behavior}. 

\subsection{Risk Assessments}

Risk assessments are mandatory for most robotics applications in industrial production within an EU regulatory context. They could be done in correspondence
with the Machinery Directive (2006/42/EC) \cite{MachineryDirective:2006} or according to the robotic harmonised standards such as ISO 12100 \cite{ISO12100:2010}. 
Risk assessment typically begins with hazard identification, estimation, and analysis of their severity, which ends with risk reduction. However, the above standards lack guidelines for assessing intelligent and collaborative robotics systems. Risk assessments are criticized in practice for being time-consuming \cite{bdiwi2022towards}.

Failure mode effect analysis (FMEA), well established in the automotive industry \cite{banduka2016integrated}, is an engineering analysis method based on reliability theory \cite{schneider1996failure}. It assesses systems, processes, or software by evaluating potential failures, their effects and causes, and determining current controls to prevent or detect failures. 
FMEA employs a bottom-up approach to identify failure modes and causes in an easy and structured manner.

\begin{figure}[t]
	\includegraphics[width=0.9\linewidth]{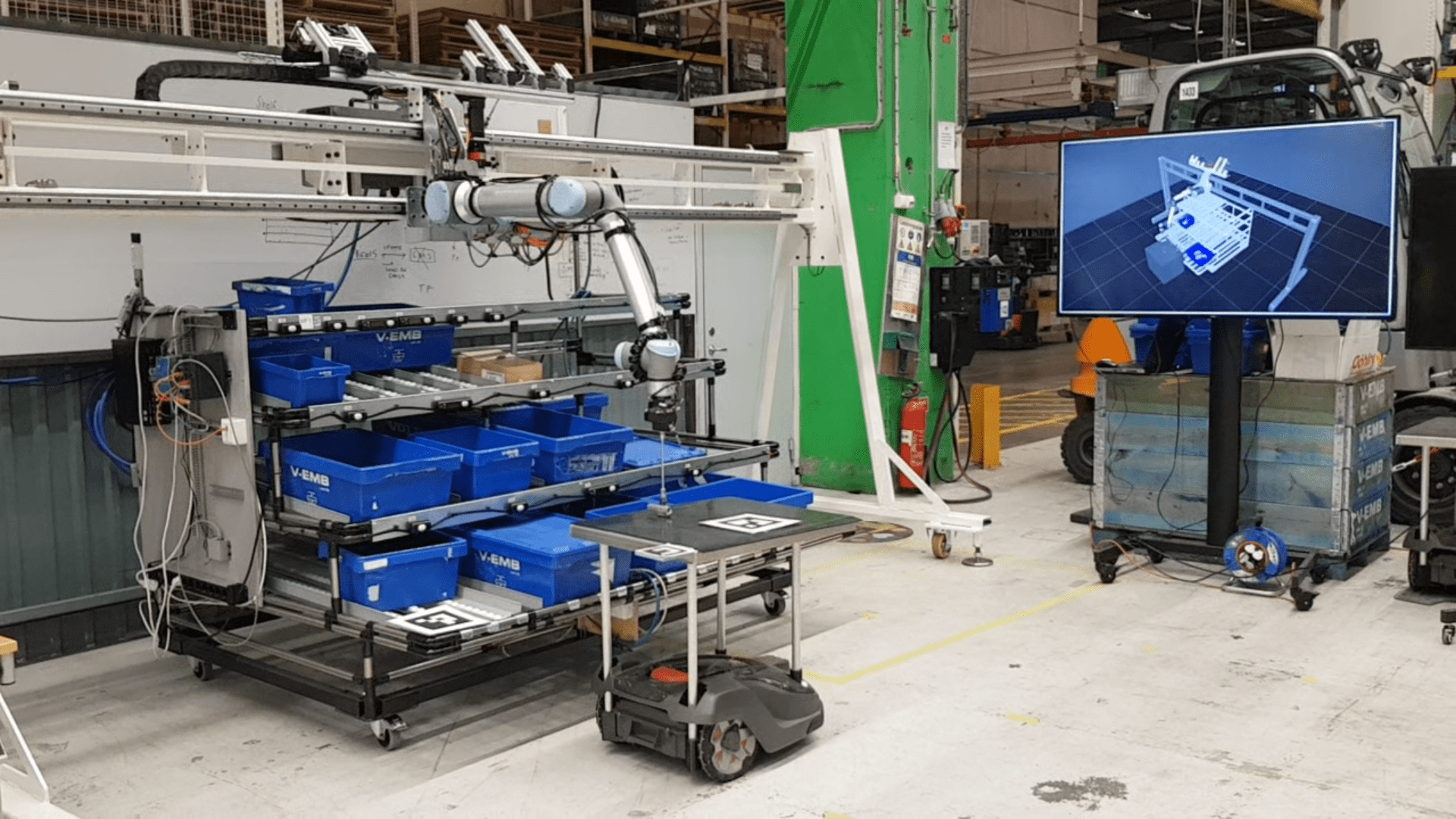}
		\caption{The collaborative robotic system (Robot in the Air).}
		\label{fig:rita}
        \vspace{-5mm}

\end{figure}

\subsection{The Collaborative Robotics Mission}
\label{subsec:usecase}

We investigate using a collaborative robotic system called Robot in the Air (RITA), an intelligent automation system for assembling kits in an assembly line.
RITA is a pick-to-light system, which is a paperless picking system with a light-directed picking \cite{battini2015comparative}. \Cref{fig:rita} shows RITA, which is composed of a gantry, a mounted robotic arm equipped with a scanner, and an autonomous trolley.
The scanner detects and localizes items to be picked from toolboxes (blue boxes in \cref{fig:rita}). 
The system should allow the operators and robots to work in a shared zone. The system can perform tasks by picking ordered items and putting them on the autonomous trolley to bring the materials to an assembly station. 
Controlling such a system involves various challenges, including but not limited to ensuring the precision and reliability of scanning and localization of items, preventing collisions between the robot and operators or surrounding equipment, quickly and adequately responding to failures, and achieving a balance between execution speed to meet assembly cycle deadlines and ensuring operator safety and equipment protection.


\section{Research Method}
\label{sec:method}
We adopted the design science method~\cite{hevner2004design} to create our model-based approach for early risk assessment.
Two research questions guided our study:

\textit{RQ1.} How can a model-based approach be designed to support risk assessments in a development-centric way?

Our goal was to investigate how we can support practitioners in robotics while doing risk assessments. As stated in the introduction, the approach relies on behavior trees, given their popularity, understandability, and support to model reactiveness.

To understand the requirements for such an approach, we conducted a workshop to understand the challenges faced by practitioners in an automotive company. Risk assessments were identified as a challenging step in robotics projects. Through follow-up meetings at the company, we discussed using behavior trees to mitigate some of the challenges, and we collaborated with the involved practitioners to develop an approach to using behavior-tree models during the risk assessment of robotics missions.

\textit{RQ2.} To what extent can our proposed approach support practitioners in performing risk assessments in a development-centric way? 

Design science involves the iterative understanding of practical problems, the development of a design artifact (i.e., our risk assessment approach), and the evaluation of the artifact~\cite{hevner2004design}.
We performed two cycles and involved practitioners from several companies.
In cycle 1, we relied on an internal evaluation with a developer who was not involved in the initial conception of the approach. In cycle 2, we included participants from external teams/companies to evaluate the approach.

\subsection{Selected Companies and Participants}
The proposed approach was developed with practitioners from an automotive manufacturer (Company A) with more than 104,000 employees, and a research and development organisation (Company B) with more than 100 employees that works closely with Company A. 

\textbf{Internal Research Team:}
Throughout this paper, we use the term ``internal participants'' to refer to the internal research team.
Two practitioners were involved during the whole study from Companies A and B.
During the work, we divided ourselves into two teams to reduce bias.

Team 1 comprised the safety expert from Company A and a researcher from academia (first and second authors).
Team 1 was involved in creating our proposed approach and applying it to the collaborative robotic mission.  

Team 2 is the industrial researcher from Company B (third author), who has technical experience with the system RITA.
In the rest of the paper, we refer to him as "the developer" of the robotic system under consideration.
Team 2 was involved in the initial internal evaluation of the process outputs. 

\textbf{Participants in cycle 2:} 
We conducted a think-aloud study to evaluate our proposed approach with external participants. The participants were chosen based on previous experience with risk analysis, availability, and existing connections. Five participants from four different companies were involved. \Cref{tbl:participants} provides an overview of the participants. P1 and P2 are from different teams in Company A, and P3 is from Company B, with no prior involvement in the study. P4 is a safety and security researcher from an independent research institute collaborating with universities, industry, and the public sector. P5 is a senior product security engineer at a medical equipment manufacturer. We use the term ``external participants'' to refer to the participants from cycle 2.

The external participants have diverse backgrounds and knowledge.
Four participants have previous experience with industrial robotics, ranging from 2 to 10 years. P1, P2, and P3 have prior experience with the system RITA under study. None of the participants had worked with behavior trees before.

\begin{table}[t]
	\caption{Overview of the external participants (cycle 2).}%
	\label{tbl:participants}
	\scalebox{0.95}{
		\begin{tabularx}{\columnwidth}{
				>{\small\raggedright}p{4mm}
				>{\small}p{22mm}
				>{\small}p{31mm}
                >{\small}p{22mm}}

			& \textbf{Role}
			& \textbf{Domain and company}
			& \textbf{Exp.~with safety analysis}
			\\
			\midrule		
			P1
			& senior researcher
			& automotive (Comp.~A)
			& 5-10 years
			\\	
			P2
			& senior researcher
			& automotive (Comp.~A)
			& 5-10 years
			\\
			P3
			& researcher
			& research and development (Comp.~B)
			& 0 years
			\\
			P4
			& safety and security researcher
			& research and development (Comp.~C)
			& 2-5 years
			\\	
			 P5
			 & product security engineer
			 & medical (Comp.~D)
			 & 5-10 years
			\\
			\bottomrule
	\end{tabularx}
	}
		\vspace{-5mm}
\end{table} 

\subsection{Understanding the Environment}
\label{subsec:problem}

The initial phase involved understanding the challenges and requirements for risk assessment. We started by conducting a workshop.
The first author held a workshop with the internal participants following Säfsten and Gustavsson guidelines~\cite{safsten2020research}. During the session, behavior trees were explained, and then we discussed challenges faced in robotics projects.
The researcher took notes on a whiteboard and paper, and later analysed them. Based on this analysis, we derived a list of requirements to overcome the challenges faced during risk assessments (cf.~Section~\ref{subsec:challenges}).

\begin{figure*}[t]
	\begin{center}
		\includegraphics[
		width=0.8\linewidth
		]{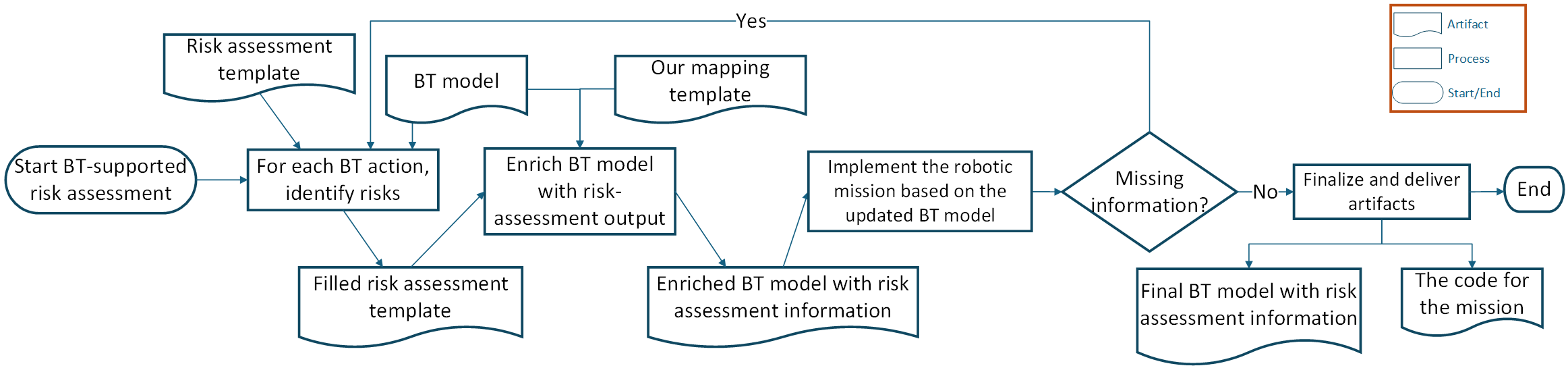}
	\end{center}
	\caption[Our process]{Our model-based approach for supporting risk assessment with BTs.}%
	\label{fig:approach}
	
	 \vspace{-5mm}
\end{figure*}

\subsection{Developing the Approach}
\label{subsec:develop}

We developed the model-based approach based on the extracted requirements from the workshop.
The researcher (first author) visited the company multiple times for a month to develop the approach collaboratively with the safety expert. The researcher took notes during the sessions, and the process of developing the approach was iterative.

\subsection{Evaluation}
\label{subsec:evaluate}

\textbf{Internal Evaluation. }
The goal of the internal evaluation in cycle 1 was to test whether an enriched behavior-tree model would support developers in understanding safety concerns and transferring them into the implementation of the robotic mission.
The developer, Team 2, was involved in this evaluation. Team 1 handed the developer the enriched behavior-tree model with the FMEA results to implement (code) the mission. 
We chose an existing behavior-tree tool to visualise the enriched behavior-tree model, called Groot.
The developer provided feedback to Team 1 to improve the level of information annotated in the behavior-tree model.
The feedback led to the iterative refinement of our risk assessment approach.

\textbf{External Evaluation. } 
In cycle 2, we conducted a think-aloud study \cite{shull2008guide} to evaluate the usefulness of using a behavior-tree model for the risk assessments of robotics missions, specifically in identifying risks, transferring the outputs into the implementation phase, and visualizing the outputs.

We organised individual think-aloud sessions with the external participants, held by the same researcher (first author). Each session was an hour long and was recorded for analysis. It started with a brief presentation by the researcher to introduce the behavior-tree model and the system RITA. 
Informed consent was obtained from each participant.
The participants were given a document describing a pick-and-place mission for the robotic arm and a behavior-tree model of the mission. The document included two tasks: (1) conduct an FMEA for the place part of the mission following our model-based approach, (2) transfer the outputs of the FMEA into a behavior-tree tool to help with the visualisation of the outputs and the outputs' transfer into the implementation phase. We used Groot to facilitate the visualisation and mapping of the outputs of the FMEA into the behavior-tree model.
We also provided an Excel file to conduct the FMEA with a pre-filled example for the pick part of the mission to guide the participants.
All documents are available in our online appendix \cite{appendix:online}.

At the end of the sessions, we provided a short survey of three parts. The first part was the raw NASA Task Load Index (RTLX) \cite{hart2006nasa} to assess the overall workload of our approach. The RTLX used a scale from 1 to 10 for each subscale, and no weighting was used for the dimensions. The second part concerned reflections on the usefulness and the most favourable and the least favourable aspects of using the behavior-tree model during the given tasks. The last part was about the future usage of the whole approach, the needed improvements, and usage for documentation.

To transcribe the recordings, we used otter.ai \cite{otterai2025} to generate initial transcripts, then we listened again to the recordings and corrected mistakes. We conducted a bottom-up thematic analysis by finding codes and themes that emerged from the data~\cite{cruzes2011recommended}. Two of the involved researchers conducted a data analysis session to derive the final themes. 

\subsection{Threats to Validity}
\label{subsec:threat}
\textbf{Internal.} 
A potential threat to internal validity is the bias of the researcher in the qualitative analysis of the think-aloud data. To mitigate that, we had one researcher systematically analyse the transcripts of the recordings and extract codes and themes. Another researcher was involved in a data analysis session to check the analysis and derive extracted themes. 

Another threat is the transcription accuracy, which affects the data interpretation. To mitigate that, we conducted a manual transcript verification and cleaning process by listening to the original recordings against the transcripts. We corrected transcription errors and the wrong assignment of speaker roles.

Another threat is recording the think-aloud sessions and using an automatic transcription tool. We informed the participants about recording the sessions, their right to withdraw, and how their data would be processed. We received informed consent from all participants. The videos were stored locally, and we limited access to them to only the two researchers who analysed the data. We only used the automatic transcription tool to transcribe the audio. We picked a tool that is System and Organisation Controls 2 (SOC 2) type 2 certified, and adheres to the European General Data Protection Regulation (GDPR), ensuring data security and privacy. 

\textbf{External.}
A major threat to the generalizability of our findings is the external participant selection, which might influence the results. Evaluating our approach required a certain background and expertise with cyber-physical systems, which required us to reach out to our network. We ensured the selection of participants from multiple companies in different domains. It allowed us to collect and analyse different perspectives on the approach.


\section{Supporting Risk Assessments (RQ1)}
\label{sec:results}

\subsection{Requirements for a Risk Assessment Approach}
\label{subsec:challenges}
The requirements were derived from the discussed challenges during the workshop and in follow-up meetings in cycle 1.

\textit{Req 1:} The approach shall support early risk assessment with subsequent iterations:  Research recommends using risk assessments at the early-design stages of projects and updating them iteratively; however, risk assessments were done later in most projects. Company A often uses failure mode and effects analysis (FMEA). FMEA was based on historical data and experience from previous projects. There is a lack of understanding of the components involved in robotics missions.

\textit{Req 2:} The approach shall have a minimal dependence on previous data and experience: With the growth of industry 5.0 \cite{toncian2024leveraging} and the introduction of new intelligent and collaborative robotics systems, it is challenging to have previous data or expertise that safety analysts can rely on. The approach should be applicable to greenfield development.

\textit{Req 3:} The approach shall support developers in understanding relevant safety concerns when implementing a mission: It was emphasised that ensuring that the output of risk assessment is adequately considered when implementing robotics missions is challenging and time-consuming. Therefore, our approach should make it easy to track the outputs of risk assessments and reflect them in the implementation code of robotic missions.

\textit{Req 4:} The approach shall support visualisation: Current approaches make it challenging to obtain an overview of the robotics missions' components and understand the main flow of the missions while doing risk assessments. There is a need to provide visualisation and a better understanding of the mission behavior \cite{hutchinson2014model}.

\textit{Req 5:} The approach shall support lightweight documentation in an easily accessible way: Previous research has found that ensuring traceability and collaboration becomes difficult when multiple organisational units and multiple tools are involved~\cite{wohlrab2020collaborative}. It would be undesirable if risk assessment findings are stored in a separate document repository, which might lead to information loss when implementing (coding) the missions. 
Instead, the approach should bridge the gap between developers and risk assessment experts by making risk assessment documentation easily accessible.

\begin{table*}
	\caption{\label{tbl:risktemplate} An excerpt of FMEA results. The mapping field shows the template for transferring FMEA outputs into the behavior-tree tool.}%
	\footnotesize
	\centering
	\scalebox{0.8}{
	\begin{tabularx}{\linewidth}{
			>{\small\raggedright}p{19mm}
			>{\small}p{38mm}|
			>{\small}p{85mm}|
			>{\small}p{5mm}}
			
		& \textbf{FMEA} 
		
		& \textbf{example}
		& \textbf{mapping}
		\\
		\midrule		
		
		& part	
		& pick &
		\\		
		\midrule
		& pre-requirement
		
		& blue box is scanned; the right gripper is attached; an order for picking items is sent by the operator/system
		& "\_description"
		\\
		\midrule
		characteristics of failures		
		& potential failure mode \newline  potential effect(s) of failure \newline potential causes of failure	
		& (1) item not picked; (2) item dropped \newline (1 and 2) process delay (performance affected) \newline (1 and 2) grasping point not accurate or gripper performance deteriorated;
		& "\_failureIf" \newline "\_description" \newline "\_description"

		\\
		\midrule
		future controls
		& controls detection \newline controls prevention \newline recommended action 
		& (1 and 2) to detect, enable the force/torque  sensor to check if the item in the gripper, or operator detects failure; \newline (1 and 2) NA \newline (1 and 2) after \# of attempts stop execution and notify operator
		& "\_failureIf" \newline "\_description" \newline "\_onFailure" 
		
		\\
		\bottomrule
	\end{tabularx}	}
    	\vspace{-5mm}
\end{table*}

\subsection{The Approach of Using Behavior Trees in Risk Assessment}
\label{subsec:approach}

Our approach integrates \bts into the risk assessment approach.
The reason to use \bts is twofold:
By using behavior trees in the process, we have two levels of information needed in risk assessments: the actions involved in a mission (execution nodes) and the decision structure showing how actions lead to specific outcomes (control nodes). Having this high level of abstraction of robotics missions with the graphical representation of the decision-making process might facilitate better identification of actions' requirements and possible failures. 
Behavior trees might also be used in the implementation and execution phases of robotics projects. Investing time in the design phase to conduct risk assessments using behavior trees could be highly effective, feeding the outputs into the subsequent phases of the project.

\Cref{fig:approach} presents our proposed approach for model-based risk assessment using \bts. 
The approach relies on a preliminary behavior-tree model.
In our case, the robotic mission is a collaborative system that prepares a full kit for an operator using the gantry and the robotic arm in a safe and timely manner. 
The full results of our example risk assessment can be accessed in our online appendix \cite{appendix:online}.  

\begin{figure}[t]
	\begin{center}
		\includegraphics[
		width=\linewidth
		]{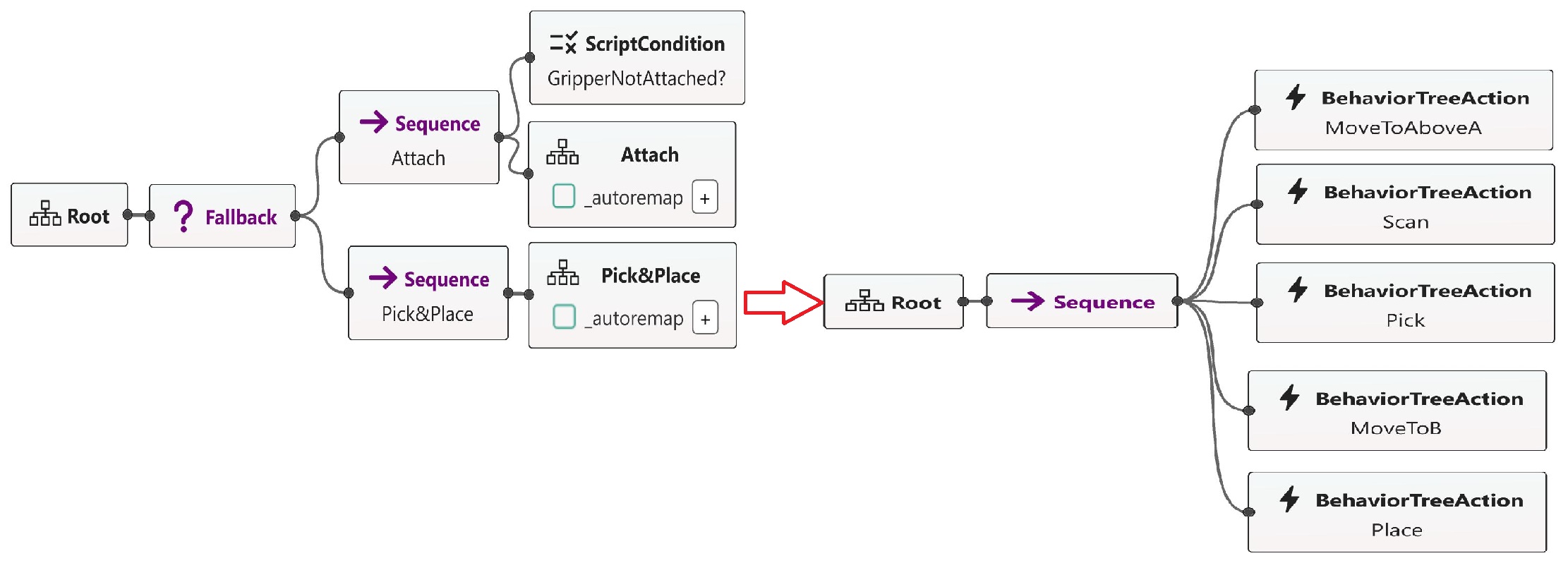}
	\end{center}
	\caption[Our mission]{The BT model representing part of the mission using Groot. On the right is the expanded subtree for Pick\&Place.}%
	\label{fig:mission}
	
	\vspace{-5mm}
\end{figure}

\Cref{fig:mission} shows our initial behavior-tree model for the mission. The robot has two tasks: attaching the right gripper (\lstinline!Attach! sequence) and picking and placing items (\lstinline!Pick&Place! sequence). The robot starts by checking if the right gripper is attached (\lstinline!GripperNotAttached?! condition) and executes the \lstinline!Attach! subtree if needed. Otherwise, the robot moves to the \lstinline!Pick&Place! sequence, where it moves the arm above the desired blue box position (\lstinline!MoveToAboveA!), scans and identifies the desired item (\lstinline!Scan!), picks the item (\lstinline!Pick!), then moves to the autonomous-robot station  (\lstinline!MoveToB!) and places the item (\lstinline!Place!).

The next step is to choose the preferred template for risk assessment. We relied on a process-FMEA template typically used at Company A. The risk assessment starts by taking each action in the behavior-tree model, like \lstinline!Pick!, and identifying pre-requirements, potential failures, the effects of a failure, its causes, control prevention, control detection, and recommended actions to mitigate the failure.
When applying FMEA, assigning severity, occurrence, and detection numbers to calculate RPN can be postponed. It is often challenging to assign the numbers at the early stages of the project without real data from studies to feed the numbers, which is a known shortcoming of FMEA \cite{liu2013risk}. 

\Cref{tbl:risktemplate} shows an excerpt of the risk assessments' outputs when applying it to the \lstinline!Pick! action performed by Team 1. 
Two possible failures were identified: (1) the desired item is not picked, or (2) the item is dropped after picking. In this case, there are similarities in the characteristics of the failures and future controls. 

After having the risk assessment's outputs, the next step is to map them in the behavior trees. In general, most behavior-tree tools offer the possibility to define when nodes fail and succeed, and what to do in case of failure or success. We recommend using the tool Groot, which allows setting the success and failure of a node and provides a "description" field. Using the mapping in \cref{tbl:risktemplate}, the outputs of FMEA can be added to the behavior-tree model (see \cref{fig:mission} for the model).

\begin{lstlisting}[style=xmlstyle, caption={\label{ls:xmlnotation}An excerpt of \BTCPP XML notation for Pick action after mapping the risk assessment outputs into behavior trees.}]
	<BehaviorTreeAction name="Pick"
	action_name="bt_action_service"
	command="pick"
	_description="Potential effect of failure is 
	process delay;
	potential causes of failure are grasping points not 
	accurate 
	or gripper performance deteriorated;
	Remember to enable the force/torque sensor to 
    detect if the item is in the gripper"
	_successIf="ItemPicked"
	_failureIf="NoItemPicked: to detect enable 
	the force/torque sensor if the item is 
    in the gripper;
\end{lstlisting}

\Cref{ls:xmlnotation} shows an excerpt of \BTCPP XML-like language that is auto-generated by Groot after saving the behavior-tree model. The XML-like file is usually imported as an external file in the implementation code to provide the syntax tree. Mapping the failures into the behavior-tree tool is straightforward. The failures can be mapped into the "\_failureIf" field with the control detection information. The rest of the characteristics of failures can be mapped into the "\_description" field. The recommended mitigation in case of failure is mapped into the "\_onFailure" field.

Finally, the final step is providing the mapped outputs in behavior trees to the developer to implement (code) the action nodes. Having the potential risks and related information should guide the developers during implementation. Risk assessment is an iterative process. Other risks might arise during the project implementation and evaluation phases, so we recommend keeping the behavior-tree model updated.


\section{Evaluation Results (RQ2)}
\label{sec:evaluationresults}

\begin{figure}
	\begin{center}
		\includegraphics[
		width=\linewidth,
		clip,
		trim=5 5 5 5
		]{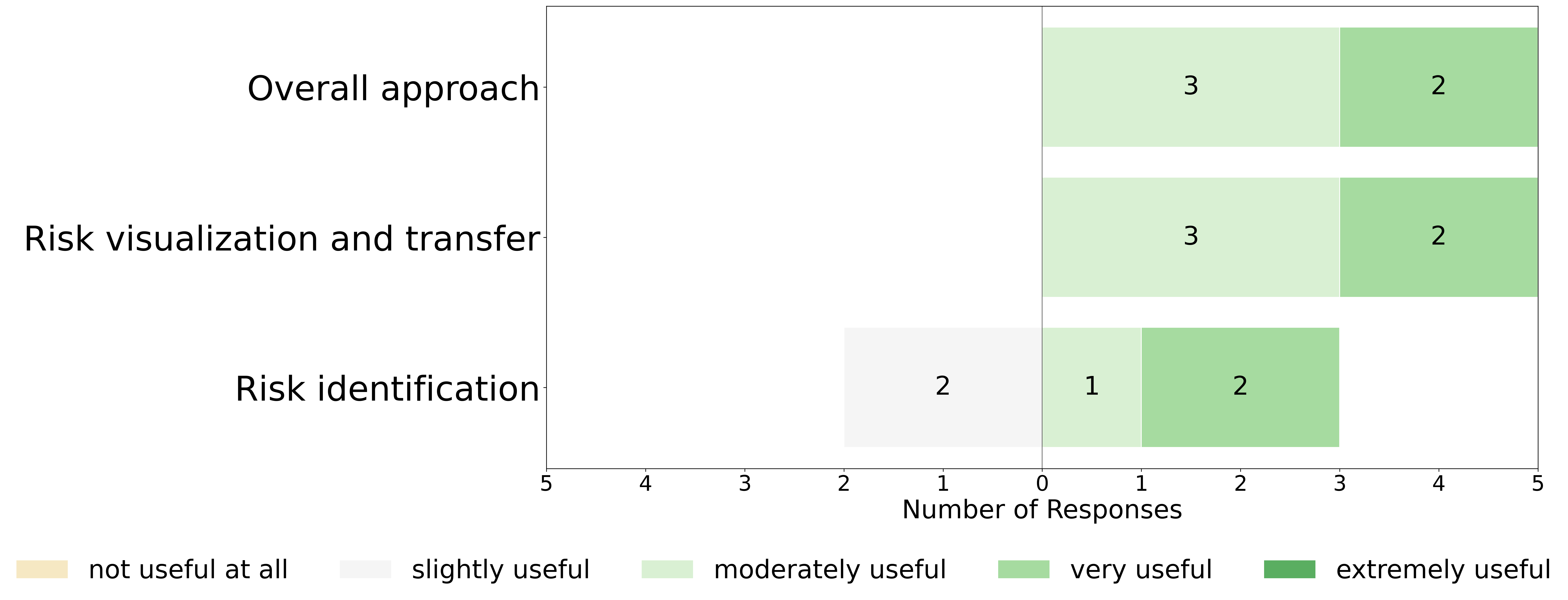}
	\end{center}
	\caption[Response Rate]{Responses on the usefulness of our approach.}%
	\label{fig:usefulness-response}
	
		\vspace{-5mm}
\end{figure}

\Cref{fig:usefulness-response} presents the results about the usefulness of our approach reported by the external participants in cycle 2. It can be seen that the participants found the overall approach of using behavior-tree models in risk assessment very useful. The usefulness was considered higher for the overall approach and for the risk visualization and transfer than for the risk identification. 

\begin{figure}[t]
	\begin{center}
		\includegraphics[
		width=0.6\linewidth,
		clip,
		trim=6 6 6 6
		]{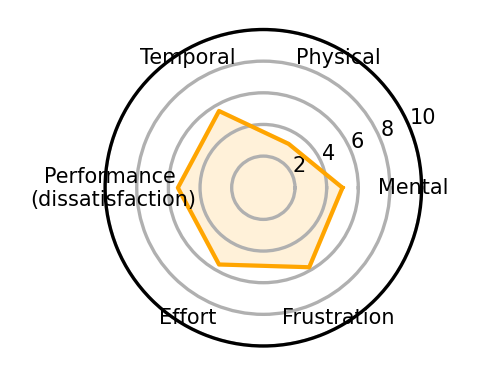}
	\end{center}
	\caption[RTLX Response]{Perceived workload for using our approach.}%
	\label{fig:rtlx-response}
	
		\vspace{-5mm}
\end{figure}

\Cref{fig:rtlx-response} shows an overview of the perceived workload by the external participants after using the behavior-tree-based approach for risk assessment. The dimensions are ranked according to the dimensions of NASA Raw TLX~\cite{hart2006nasa}, where higher numbers correspond to a higher workload. Overall, the participants experienced a moderate level of perceived workload (average 5.1 on a 1–10 scale), and their experiences were consistent (standard deviation 1.6).

\textbf{Mental Model for Risk Identification.}
All participants found it helpful to have the behavior-tree model as a starting point to familiarize themselves with the robotic mission. The behavior tree provided an entry point for thinking about what can go wrong in the robotic mission. 
It was reported that organising actions and decisions in a tree-like manner made the potential failure points clearer. 

While identifying risks during the FMEA, we noticed two patterns in participants. P1 and P2, who had previous experience with the system RITA, did not use the behavior-tree model as intensely and found the provided model to be somewhat simplistic. At the same time,  one participant with prior experience with RITA (P3) and two participants with no prior experience with the system (P4 and P5) appreciated having the model while eliciting the needed information for FMEA. It was stated that the behavior-tree model helped them build their mental model of the mission steps:

\textit{P4: I can go back to the figure when I am stuck, because sometimes I can infer what can go wrong. [...] You need the visual sometimes to just remember things. [...] It's like any supporting artifact, even if you know it, sometimes you need some kind of confirmation, so you go back and try to read it one more time while you are thinking.}

We found that it is important to remind stakeholders using the behavior tree that it is a living model that should evolve as risks are identified, e.g., by adding components to overcome the risks.
For P3 and P4, having the behavior tree might have restricted their brainstorming, and we reminded them that they could add things and that the model was an initial one.

\begin{finding}
	Using behavior trees at the early stages of risk assessment enables the representation of both the system's intended nominal behavior and potential failure points within a single, structured model, which can help practitioners create a mental model of the mission and associated risks.
\end{finding}

\textbf{Documentation and Version Control.} 
All participants stated that they saw benefits in using the behavior-tree model in the risk assessment approach for documentation, even if it is not ultimately implemented as executable behavior trees in the project.
However, P5 expressed concerns regarding tracking changes and keeping information up-to-date across the different locations, especially in a regulated industry. 

During the think-aloud sessions, P3, P4, and P5 did not initially transfer all the information from the behavior tree. 
This shows the possibility of information loss and the need for automation. P4 and P5 expressed the need to automate the mapping of risk assessment outputs into the behavior-tree tool.

\textit{P5: Now you're duplicating information that's going to be in multiple places. [...] If I were in a completely unregulated field, that's not really a big deal. [...] As I move into more complex systems and regulated industries, this can be a challenge if you don't have automation in place. [...] From a long-term maintenance perspective, it becomes a challenge in the regulated industries.}

Our iterative approach suggests updating and remodelling the behavior trees as failures are identified throughout the project. P3 expressed the need for a version control system to track changes since the approach is iterative. It was expressed that each behavior tree and its associated FMEA evolve in parallel with each iteration of addressing identified failures, necessitating systematic tracking of changes to maintain consistency and traceability.

\begin{finding}
	The behavior-tree model provides a promising model for holding and documenting risk assessment information. However, to maintain consistency and traceability, information transfer to existing tools needs to be automated, and a version control system needs to be established.
\end{finding}

\textbf{Granularity of the Model.}
In the think-aloud tasks, we used a low level of granularity (fewer details) to provide a clear and easy-to-follow behavior tree.
P1, P2, P3 and P4 desired a high-granularity behavior-tree model (more details) to reduce the need for assumptions. 

\textit{P4: I was missing some more information to not make many assumptions. [...] 
I feel it would have been easier with a more comprehensive view of the system behavior.} 

P4 also noted that with a higher level of granularity, understanding the model might demand extra time, and highly complex and larger behavior-tree models might cause confusion. 
We acknowledge that it is hard to find the balance between high- and low-granularity models to provide clear robotics missions without being simplistic. In addition, providing a detailed behavior-tree model might be challenging with new systems and might skew the brainstorming of potential failures. More research is needed to determine the balance in the granularity of the provided model.

\begin{finding}
	The behavior-tree model’s level of granularity is challenging to balance to provide a clear and understandable mission without biasing the risk assessment brainstorming. 
\end{finding}

\textbf{Tooling for Behavior Trees.} 
There were ambivalent experiences with the used behavior-tree tool (Groot).
Groot is widely used for behavior-tree modeling, and all participants appreciated the GUI and visualization. However, the tool was not appreciated for transferring and storing the identified failure information. The tool is not designed specifically to hold risk assessment information, which may have hindered the user experience.

\textit{P4: I just found the tool to not be user-friendly, and that slightly affected my overall perception of using the BTs for these tasks negatively.}

In general, behavior trees are appropriate models to identify risks and potential failures, as they require defining when nodes succeed or fail. Thus, most existing tools will have fields to map this information.
We found Groot to be beneficial, partially because of the available data fields for the identified failure risks and recommended actions. One challenging aspect was mapping the rest of the outputs (pre-requirements, effect and cause of failure, and controls detection and prevention). Since Groot had a free-text field for description for each node, we mapped the information into it. However, it is not necessarily an optimal way to present the information. 
P1, P3, and P4 highlighted that connecting information recorded in the free-text field with that entered into structured fields for specific failures might lead to traceability issues, as all this information pertains to the same failure event.

\textit{P3: It's a bit hard when you have this free text... I need some way to make it clear which failure is related to the information in the description.}

All participants expressed the need for better tooling that connects risk assessment artifacts to code in a smoother way. P2, P3, and P4 stated that it would be beneficial to conduct risk assessment directly in the behavior-tree tool---partially because it would save time and cognitive effort, and because it would make it easier to connect the identified information in the risk assessment directly to the code.
All participants found switching between tabs problematic. When the outputs of the risk assessments were transferred into the behavior-tree tool, participants reported losing the overview of information compared to the Excel worksheet. One participant preferred entering all information related to a single failure at once.

\textit{P3: Maybe you could add an FMEA tab here, and then you have all this, exactly these headlines.. then you can say that you follow the FMEA... an added value here is the connection to the source code.}

We believe the aforementioned positive aspects and limitations expressed could be used as requirements to extend existing behavior-tree tools to support the early identification of risk information that influences the code.

\begin{finding}
	The behavior tree tool Groot could naturally model information about potential failures and recommended actions from FMEA, but it lacks optimal support for other types of information. In addition, the choice of the behavior-tree tool affects the process of transferring and visualizing the outputs of risk assessment. This highlights an opportunity for the modelling community to improve existing tools.
\end{finding}

\textbf{The Gap Between Risk Assessment and Implementation.}
In the internal evaluation in cycle 1, the developer used the enriched behavior-tree model in Groot to implement (code) the robotic mission. 
The enriched behavior-tree model conveyed nominal behaviors, expected sequences, possible failure modes, and contingency actions for various scenarios—all of which stem directly from the risk assessment.
This approach ensured that the developer had direct access to comprehensive information about both standard and failure scenarios, removing ambiguity and reducing the need for interpretative guesswork. 

The external participants in cycle 2 did not perform the task of implementing the resulting behavior trees. However, P5 anticipated the potential of having the enriched behavior-tree model with risk assessment information to improve the development process---a perspective substantiated by the developer's practical experience. In addition, P2 and P3 noted that they valued the ability to connect and transfer risk information to the code for later stages in the projects. 

\textit{P5: It always comes down to how experienced your engineers and team are. Having the right assumptions or mental model going in. When I hire new college grads or software engineers, I can't expect them to know all the failure modes. So, it can be a lot of a higher value to have the behavior tree with risks right in front of them.}

\begin{finding}
	We can ensure alignment between high-level risk assessments and low-level implementation (code) by leveraging the behavior-tree model as an information bridge for risk assessment outputs. 
\end{finding}

\section{Related Work}
\label{sec:relatedWork}
Bdiwi et al. \cite{bdiwi2022towards} and Abdulkhaleq and Wagner \cite{abdulkhaleq2013integrating} proposed integrating state machines into risk assessments, which are another type of behavioral model. Bdiwi et al used state machines for modelling risk mitigation strategies, and Abdulkhaleq and Wagner used them to provide appropriate diagrammatic notations to represent the safety control structure in systems theoretic process analysis (STPA).
The aforementioned research highlights the importance of a clear visual behavior model illustrating the relationships between mission components and control actions to identify potential risks. Unlike our work, none of the previous works explored integrating the behavior-tree model into risk assessment approaches.

Previous research~\cite{kokotinis2024alpha,castano2019safe} shows the potential of behavior trees for modelling and executing mitigation strategies based on system constraints such as safety. In the framework proposed by Castano and Xu \cite{castano2019safe}, behaviour trees were used to model risk mitigation strategies by detecting failures and reacting to mitigate them. Although risk assessments are mentioned as an important component of their framework, they do not mention the risk assessment used or the process for mapping the risk assessment's outputs.
Our work builds on this potential and advocates for incorporating the behavior-tree model into the risk assessment approach from the early design phase of robotics projects, especially in industry, to improve the transferability and communication between stakeholders.


\section{Conclusion and Future Work}
\label{sec:conclusion}

This work is the first step to support risk assessments using the behavior-tree model. We developed and provided a model-based approach for using behavior trees in risk assessments with the support of industrial practitioners. Our approach uses the behavior-tree model to support risk identification, visualization, and transfer of risk assessments' outputs. We evaluated the approach with practitioners from different companies. Our findings highlighted the potential of the behavior-tree model in supporting practitioners in forming a better understanding of the robotic mission to identify potential failure points in the early stages of the project. They also identified the potential of aligning code implementation with the risk assessment outputs and forming lightweight documentation. However, further development of our approach is needed to automate the information transfer of risk assessment information to the implementation, select the granularity of the model, and provide better tooling.

In future work, we want to measure the improvement in identifying risks when using behavior trees compared to traditional risk assessment processes. Finally, we want to improve existing tooling to support information transfer.

    \section*{Acknowledgment}
This work was partially supported by the Wallenberg AI, Autonomous Systems and Software Program (WASP) funded by the Knut and Alice Wallenberg Foundation.

\bibliographystyle{IEEEtran}
\bibliography{main.bib}

\begin{thebibliography}{10}
\providecommand{\url}[1]{#1}
\csname url@samestyle\endcsname
\providecommand{\newblock}{\relax}
\providecommand{\bibinfo}[2]{#2}
\providecommand{\BIBentrySTDinterwordspacing}{\spaceskip=0pt\relax}
\providecommand{\BIBentryALTinterwordstretchfactor}{4}
\providecommand{\BIBentryALTinterwordspacing}{\spaceskip=\fontdimen2\font plus
\BIBentryALTinterwordstretchfactor\fontdimen3\font minus
  \fontdimen4\font\relax}
\providecommand{\BIBforeignlanguage}[2]{{%
\expandafter\ifx\csname l@#1\endcsname\relax
\typeout{** WARNING: IEEEtran.bst: No hyphenation pattern has been}%
\typeout{** loaded for the language `#1'. Using the pattern for}%
\typeout{** the default language instead.}%
\else
\language=\csname l@#1\endcsname
\fi
#2}}
\providecommand{\BIBdecl}{\relax}
\BIBdecl

\bibitem{garcia2020robotics}
S.~Garc{\'\i}a, D.~Str{\"u}ber, D.~Brugali, T.~Berger, and P.~Pelliccione,
  ``Robotics software engineering: A perspective from the service robotics
  domain,'' in \emph{ESEC/FSE}, 2020, pp. 593--604.

\bibitem{guiochet2017safety}
J.~Guiochet, M.~Machin, and H.~Waeselynck, ``Safety-critical advanced robots: A
  survey,'' \emph{Robotics and Autonomous Systems}, vol.~94, pp. 43--52, 2017.

\bibitem{tom2013exploration}
E.~Tom, A.~Aurum, and R.~Vidgen, ``An exploration of technical debt,''
  \emph{Journal of Systems and Software}, vol.~86, no.~6, pp. 1498--1516, 2013.

\bibitem{foidl2019technical}
H.~Foidl, M.~Felderer, and S.~Biffl, ``Technical debt in data-intensive
  software systems,'' in \emph{SEAA}.\hskip 1em plus 0.5em minus 0.4em\relax
  IEEE, 2019, pp. 338--341.

\bibitem{curtis2012risk}
P.~Curtis, M.~Carey, C.~of~Sponsoring Organizations of~the Treadway~Commission
  \emph{et~al.}, ``Risk assessment in practice,'' 2012.

\bibitem{marzinotto2014towards}
A.~Marzinotto, M.~Colledanchise, C.~Smith, and P.~{\"O}gren, ``Towards a
  unified behavior trees framework for robot control,'' in \emph{ICRA}, 2014.

\bibitem{colledanchise2018behavior}
M.~Colledanchise and P.~{\"O}gren, \emph{Behavior trees in robotics and AI: An
  introduction}.\hskip 1em plus 0.5em minus 0.4em\relax CRC Press, 2018.

\bibitem{appendix:online}
``Online appendix,'' https://github.com/RazanGhzouli/2025-BT-risk-assessment,
  2025.

\bibitem{ghzouli2023behavior}
R.~Ghzouli, T.~Berger, E.~B. Johnsen, A.~Wasowski, and S.~Dragule, ``Behavior
  trees and state machines in robotics applications,'' \emph{IEEE Transactions
  on Software Engineering}, vol.~49, no.~9, 2023.

\bibitem{MachineryDirective:2006}
\emph{Directive 2006/42/EC of the European Parliament and of the Council on
  Machinery, and Amending Directive 95/16/EC (recast)}, European Parliament and
  Council of the European Union, Brussels, May 2006, official Journal of the
  European Union, L 157, 9 June 2006, pp. 24–86.

\bibitem{ISO12100:2010}
\emph{Safety of machinery – General principles for design – Risk assessment
  and risk reduction}, International Organization for Standardization, Geneva,
  Switzerland, November 2010.

\bibitem{bdiwi2022towards}
M.~Bdiwi, I.~Al~Naser, J.~Halim, S.~Bauer, P.~Eichler, and S.~Ihlenfeldt,
  ``Towards safety4. 0: A novel approach for flexible human-robot-interaction
  based on safety-related dynamic finite-state machine with multilayer
  operation modes,'' \emph{Frontiers in Robotics and AI}, vol.~9, 2022.

\bibitem{banduka2016integrated}
N.~Banduka, I.~Veza, and B.~Bili{\'c}, ``An integrated lean approach to process
  failure mode and effect analysis (pfmea): A case study from automotive
  industry,'' vol.~11, no.~4, 2016.

\bibitem{schneider1996failure}
H.~Schneider, ``Failure mode and effect analysis: Fmea from theory to
  execution,'' 1996.

\bibitem{battini2015comparative}
D.~Battini, M.~Calzavara, A.~Persona, and F.~Sgarbossa, ``A comparative
  analysis of different paperless picking systems,'' \emph{Industrial
  Management \& Data Systems}, vol. 115, no.~3, pp. 483--503, 2015.

\bibitem{hevner2004design}
A.~R. Hevner, S.~T. March, J.~Park, and S.~Ram, ``Design science in information
  systems research,'' \emph{MIS quarterly}, pp. 75--105, 2004.

\bibitem{safsten2020research}
K.~S{\"a}fsten and M.~Gustavsson, ``Research methodology: for engineers and
  other problem-solvers,'' 2020.

\bibitem{shull2008guide}
F.~Shull, J.~Singer, and D.~I. Sj{\o}berg, \emph{Guide to advanced empirical
  software engineering}.\hskip 1em plus 0.5em minus 0.4em\relax Springer, 2008,
  vol.~93.

\bibitem{hart2006nasa}
S.~G. Hart, ``Nasa-task load index (nasa-tlx); 20 years later,'' in
  \emph{Proc.~of the human factors and ergonomics society annual meeting},
  vol.~50, no.~9.\hskip 1em plus 0.5em minus 0.4em\relax Sage publications Sage
  CA: Los Angeles, CA, 2006, pp. 904--908.

\bibitem{otterai2025}
{Otter.ai}, ``Otter.ai: Transcription service,'' 2025, accessed April 23, 2025.
  \url{https://otter.ai}.

\bibitem{cruzes2011recommended}
D.~S. Cruzes and T.~Dybå, ``Recommended steps for thematic synthesis in
  software engineering,'' in \emph{ESEM}.\hskip 1em plus 0.5em minus
  0.4em\relax IEEE, 2011, pp. 275--284.

\bibitem{toncian2024leveraging}
V.~Toncian, A.~Florea, A.~David, D.~Morariu, and R.~Cretulescu, ``Leveraging
  collaboration for industry 5.0: Needs, strategies and future directions,'' in
  \emph{Working Conference on Virtual Enterprises}.\hskip 1em plus 0.5em minus
  0.4em\relax Springer, 2024, pp. 319--335.

\bibitem{hutchinson2014model}
J.~Hutchinson, J.~Whittle, and M.~Rouncefield, ``Model-driven engineering
  practices in industry: Social, organizational and managerial factors that
  lead to success or failure,'' \emph{Science of Computer Programming},
  vol.~89, pp. 144--161, 2014.

\bibitem{wohlrab2020collaborative}
R.~Wohlrab, E.~Knauss, J.-P. Stegh{\"o}fer, S.~Maro, A.~Anjorin, and
  P.~Pelliccione, ``Collaborative traceability management: a multiple case
  study from the perspectives of organization, process, and culture,''
  \emph{Requirements Engineering}, vol.~25, no.~1, pp. 21--45, 2020.

\bibitem{liu2013risk}
H.-C. Liu, L.~Liu, and N.~Liu, ``Risk evaluation approaches in failure mode and
  effects analysis: A literature review,'' \emph{Expert systems with
  applications}, vol.~40, no.~2, pp. 828--838, 2013.

\bibitem{abdulkhaleq2013integrating}
A.~Abdulkhaleq and S.~Wagner, ``Integrating state machine analysis with
  system-theoretic process analysis,'' in \emph{Software Engineering
  2013-Workshopband}.\hskip 1em plus 0.5em minus 0.4em\relax Gesellschaft
  f{\"u}r Informatik eV, 2013, pp. 501--514.

\bibitem{kokotinis2024alpha}
G.~Kokotinis, G.~Michalos, Z.~Arkouli, and S.~Makris, ``A behavior trees-based
  architecture towards operation planning in hybrid manufacturing,''
  \emph{International Journal of Computer Integrated Manufacturing}, vol.~37,
  no.~3, pp. 324--349, 2024.

\bibitem{castano2019safe}
L.~Castano and H.~Xu, ``Safe decision making for risk mitigation of uas,'' in
  \emph{ICUAS}.\hskip 1em plus 0.5em minus 0.4em\relax IEEE, 2019, pp.
  1326--1335.

\end{thebibliography}

\end{document}